\documentclass[letterpaper, 10 pt, conference]{ieeeconf}
\IEEEoverridecommandlockouts
\usepackage{cite}
\usepackage{hyperref}
\usepackage{amsmath,amssymb,amsfonts}
\usepackage{mathtools, float}
\usepackage{graphicx,epstopdf}
\usepackage{euscript,paralist}
\usepackage{pgf,tikz}
\usepackage[utf8]{inputenc}
\usepackage{graphicx}
\usepackage{caption}
\usepackage{subcaption}
\def\BibTeX{{\rm B\kern-.05em{\sc i\kern-.025em b}\kern-.08em
		T\kern-.1667em\lower.7ex\hbox{E}\kern-.125emX}}
\newfont{\roma}{cmr10 scaled 1200}

\newcommand{\rline}  {{\mathbb R}}

\newcommand{\FORALL} {{\hbox{$\hskip 11mm \forall \;$}}}
\newcommand{\mm}    {{\hbox{\hskip 0.5pt}}}

\newcommand{\bluff} {{\hbox{\raise 15pt \hbox{\mm}}}}
\newcommand{\sbluff}{{\hbox{\raise  7pt \hbox{\mm}}}}
\renewcommand{\theequation}{{\arabic{section}.\arabic{equation}}}

\newcommand{\bbm}[1]{\left[\begin{matrix} #1 \end{matrix}\right]}

\title{Design, fabrication and control of a cable-driven parallel robot}
\author{Dhruv Sorathiya, Sarthak Sahoo and Vivek Natarajan% <-this % stops a space
	\thanks{*This work was supported by the Industrial Research and Consultancy Centre at IIT Bombay via the internship grant RI/0323-10001447-001.}% <-this % stops a space
	\thanks{The authors are with the Centre for Systems and Control Engineering, Indian Institute of Technology Bombay, Mumbai 400076, India (e-mail: {\tt\small 200100057@iitb.ac.in}, {\tt\small violinistss.muse@gmail} and {\tt\small vivek.natarajan@iitb.ac.in})}%
}
\begin{document}
\maketitle
	%------------------------------------------------------------------------------------%
	%------------------------------------------------------------------------------------%	
\begin{abstract}
In cable driven parallel robots (CDPRs), the payload is suspended using a network of cables whose length can be controlled to maneuver the payload within the workspace. Compared to rigid link robots, CDPRs provide better maneuverability due to the flexibility of the cables and consume lesser power due to the high strength-to-weight ratio of the cables. However, amongst other things, the flexibility of the cables and the fact that they can only pull (and not push) render the dynamics of CDPRs complex. Hence advanced modelling paradigms and control algorithms must be developed to fully utilize the potential of CDPRs. Furthermore, given the complex dynamics of CDPRs, the models and control algorithms proposed for them must be validated on experimental setups to ascertain their efficacy in practice.
%Experimental setups are essential for validating the models and control algorithms proposed for CDPRs, they are complex to build.
We have recently developed an elaborate experimental setup for a CDPR with three cables and validated elementary open-loop motion planning algorithms on it. In this paper, we describe several aspects of the design and fabrication of our setup, including component selection and assembly, and present our experimental results. Our setup can reproduce complex phenomenon such as the transverse vibration of the cables seen in large CDPRs and will in the future be used to model and control such phenomenon and also to validate more sophisticated motion planning algorithms.
\end{abstract} %------------------------------------------------------------------------------------%
	
	%%%%%%%**********%%%%%%%%**********%%%%%%%%%%*********%%%%%%%%%%********* %%%%%%%**********%%%%%%%%**********%%%%%%%%%%*********%%%%%%%%%%*********	
\section{Introduction} \label{sec1}  %Section 1

A cable driven parallel robot (CDPR) consists of a network of cables from which the payload is suspended. One end of each cable in the network is connected to a motor (dedicated to that cable), while the other end is connected to the payload. The length of each cable between its motor and the payload can be varied by rotating the motor shaft. To move the payload along a desired trajectory, the lengths of all the cables must be changed appropriately using their motors.  A popular example of such a robot is the `Skycam’ \cite{SkyCam}.

CDPRs have certain advantages compared to rigid link robots. In CDPRs, the payload is supported by a set of cables and motors which leads to a distribution of load and thereby minimizes the power requirement from each motor. The cables which perform the role of links in these robots are light and can be very long. This enables construction of CDPRs which can operate over large work spaces such as a stadium. Due to the light weight of the cable, CDPRs can provide rapid acceleration with lesser power consumption as compared to rigid link manipulators. For these reasons, CDPRs have found application in many areas \cite{telescope}, \cite{manufacturing}, \cite{crane}.

%They have applications in sports coverage, search and rescue operations and pick and place cranes.

While using cables instead of rigid links provides some advantages, certain properties exhibited by the cables such as flexibility (which allows for vibration of the cable), distributed weight (which can lead to sagging) and slackness (which means that the cables cannot push) render the dynamics of CDPRs complex. Existing works on modelling and control of these robots are adequate for executing typical maneuvers.
However, to fully utilize the potential of these robots advanced modelling paradigms and control algorithms must be developed. Once developed, these models and control algorithms must be validated
on experimental setups to ascertain their efficacy in practice.

In the Continuum Dynamics Control (CDC) lab\footnote{https://sites.google.com/view/viveknatarajan/cdc-lab} at IIT Bombay, we have recently built an experimental setup for a CDPR with three cables inspired by the setup in \cite{Hardware_setup_ref}. In this paper, we describe various aspects related to this setup. The paper is organized as follows: Section \ref{sec2} briefly presents a mathematical model for this setup. In Section \ref{sec3}, we describe the design and fabrication of our setup with particular emphasis on component selection and assembly. In Section \ref{sec4} we consider two simple maneuvers for the payload, design some elementary motion planning algorithms for executing these maneuvers and present the results of implementing them on our setup. %We conclude with some remarks and observations in Section \ref{sec5}.

%%%%%%%**********%%%%%%%%**********%%%%%%%%%%*********%%%%%%%%%%*********
%%%%%%%**********%%%%%%%%**********%%%%%%%%%%*********%%%%%%%%%%*********	
\section{Dynamic model for a CDPR with three cables} \label{sec2} %Section 2

Our goal is to design and fabricate a CDPR with three cables, which can be used for validating our mathematical models and control algorithms. An important step in this regard is identifying the desired specifications/ratings for the different components which will be used to build the CDPR. For this purpose we simulate (numerically in MATLAB) a relatively simple mathematical model of the CDPR to estimate the range of velocities, accelerations and tensions encountered when the CDPR is used to execute various maneuvers of the payload. Using the estimated ranges we select the motors, cables, microcontrollers and other components of the CDPR.

A simple model for a CDPR with three cables is as follows \cite{book}: For $t\geq0$,
\begin{align}
  & \hspace{10mm} \bbm{m \ddot{p}(t) + F \\ J \dot{\omega}(t) + \omega(t)\times J\omega(t)} = -H(t) T(t), \label{eq:model1}\\
  & H(t) = \bbm{s_1(t) & s_2(t) & s_3(t)\\ b_1(t) \times s_1(t) & b_2(t)\times s_2(t) & b_3(t) \times s_3(t)}, \label{eq:model2}\\
  & T_i(t) = \begin{cases}
     k(l_i(t)-l_{Ni}(t))  & \text{if } l_i(t) \geq l_{Ni}(t)\\
     0 & \text{if } l_i(t) < l_{Ni}(t) \label{eq:model3}
    \end{cases}
\end{align}
for $i\in\{1,2,3\}$. Here $m\in\rline$ and $J\in\rline^{3\times 3}$ are the mass and moment of inertia (about center of mass) of the payload, respectively, $p(t)\in\rline^{3\times 1}$ and $\omega(t)\in\rline^{3\times 1}$ denote the position and angular velocity of the payload, respectively, $F\in\rline^{3\times 1}$ is the external force due to gravity, $T(t)\in\rline^{3\times 1}$ is the tension matrix with $T_i(t)$ being the tension in the $i^{\rm th}$ cable, $s_i(t)\in\rline^{3\times 1}$ is the unit vector along the $i^{\rm th}$ cable pointing towards the payload, $b_i(t)\in\rline^{3\times 1}$ is the vector from the center of mass of the payload to the distal anchor point of the $i^{\rm th}$ cable, $l_i(t)$ and $l_{Ni}(t)$ denote the actual length and natural length of the $i^{\rm th}$ cable, respectively, $k>0$ is the cable stiffness and $H(t)\in\rline^{6\times 3}$ is the Jacobian matrix.

We simulated the model \eqref{eq:model1}-\eqref{eq:model3} numerically and used the results to select the components of our setup. We do not provide the model parameters used in our simulations due to space constraints.

%%%%%%%**********%%%%%%%%**********%%%%%%%%%%*********%%%%%%%%%%*********
%%%%%%%**********%%%%%%%%**********%%%%%%%%%%*********%%%%%%%%%%*********	
\section{ Design and fabrication of a CDPR} \label{sec3} %Section 3

We have designed and fabricated a CDPR with three cables inspired by the setup in \cite{Hardware_setup_ref}. Our setup is an under-actuated system consisting of three cables/motors (actuators) and three degrees of freedom. We will discuss the component selection for our setup as well as the fabrication of some of the components in detail below.

\subsection{Motors}
There are several types of motors such as servo motors, geared DC motors and stepper motors from which we could choose the motors for our setup. Based on the torque, speed and resolution requirements (dictated by our simulation of the model \eqref{eq:model1}-\eqref{eq:model3}), ease of operating the motor, local availability and cost effectiveness, we selected closed-loop stepper motors which can provide up to 3\,Nm holding torque and 1200\,RPM speed and whose pulses per revolution (PPR) can be chosen to be between 800 and 40000. %While a large PPR is preferable for an good accuracy, it make control

\subsection{Winch drum mechanism}
The winch drum mechanism converts the rotary motion of the motor into  linear motion of the cable. In our setup, this mechanism consists of a threaded drum which is attached to the stepper motor using a steel shaft, bearings and a flexible aluminium coupling. The radius of the drum determines the relationship between the torque/angular velocity provided by the motor and the tension/linear velocity of the cable. Hence it must  be considered while arriving at the motor specifications. Figure 1 shows a 3D printed winch drum assembled with the motor shaft.

\subsection{Payload}
We have considered two types of payloads, see Figures 3 and 4. Both payloads are cylindrical in shape and carry weights inside them. The difference between the two payloads is with regard to the distal anchor points (points of attachment) for the cables on the cylinders. While in Payload A there are three distinct anchor points located on the edge of the cylinder, in Payload B there is a single anchor point at the center to which all the three cables are attached. The dynamics associated with the two payloads are different, since the vector $b_1$, $b_2$ and $b_3$ in \eqref{eq:model2} are different for the two payloads. Hence control algorithms needed for the two payloads will also be different.

\subsection{Pulley}
A pulley is needed for guiding the cable smoothly (i.e. with low friction) from the drum to the distal anchor points on the payload. Figure 2 shows a 3D printed pulley mounted on bearings (so that the pulley can change direction).

\begin{figure}[H]
    \centering
    \begin{tabular}{cc}
        \begin{minipage}[t]{0.22\textwidth}
            \centering
            \includegraphics[width=\linewidth]{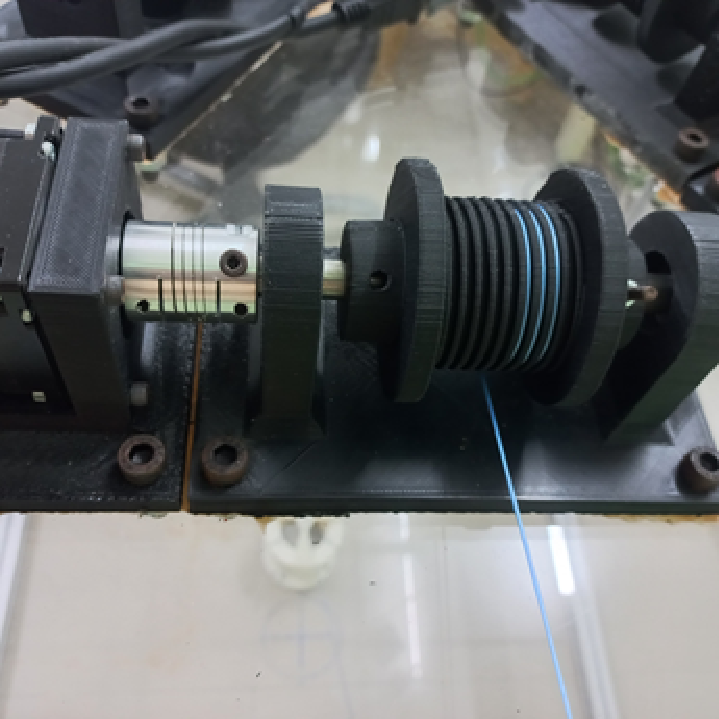}
            {\small\captionof{figure}{Winch drum mechanism}}
            \vspace{2pt} % Adjust spacing between image and caption if needed
        \end{minipage} &
        \begin{minipage}[t]{0.22\textwidth}
            \centering
            \includegraphics[width=\linewidth]{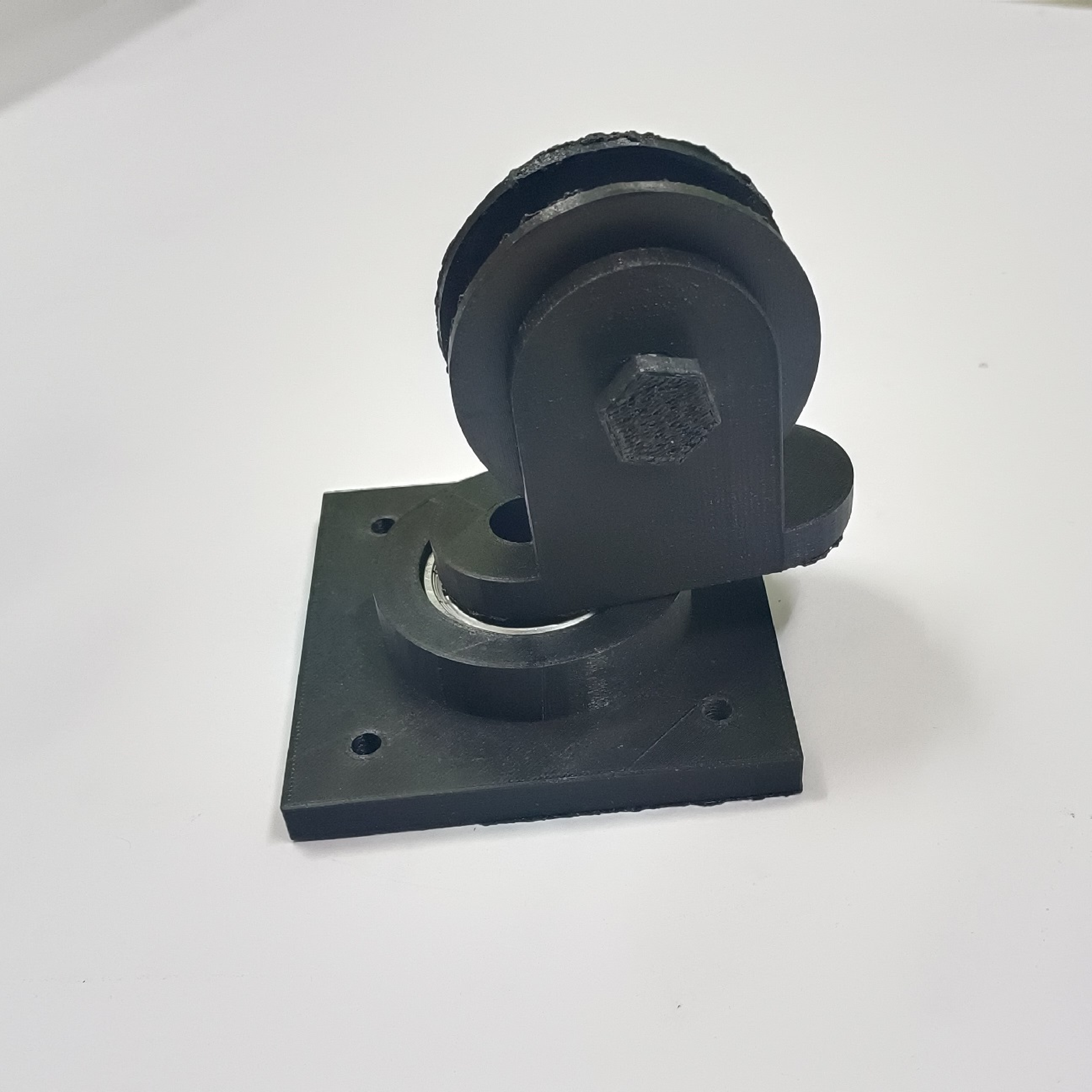}
            {\small\captionof{figure}{Pulley}}
            \vspace{2pt}
        \end{minipage} \\
        \begin{minipage}[t]{0.22\textwidth}
            \centering
            \includegraphics[width=\linewidth]{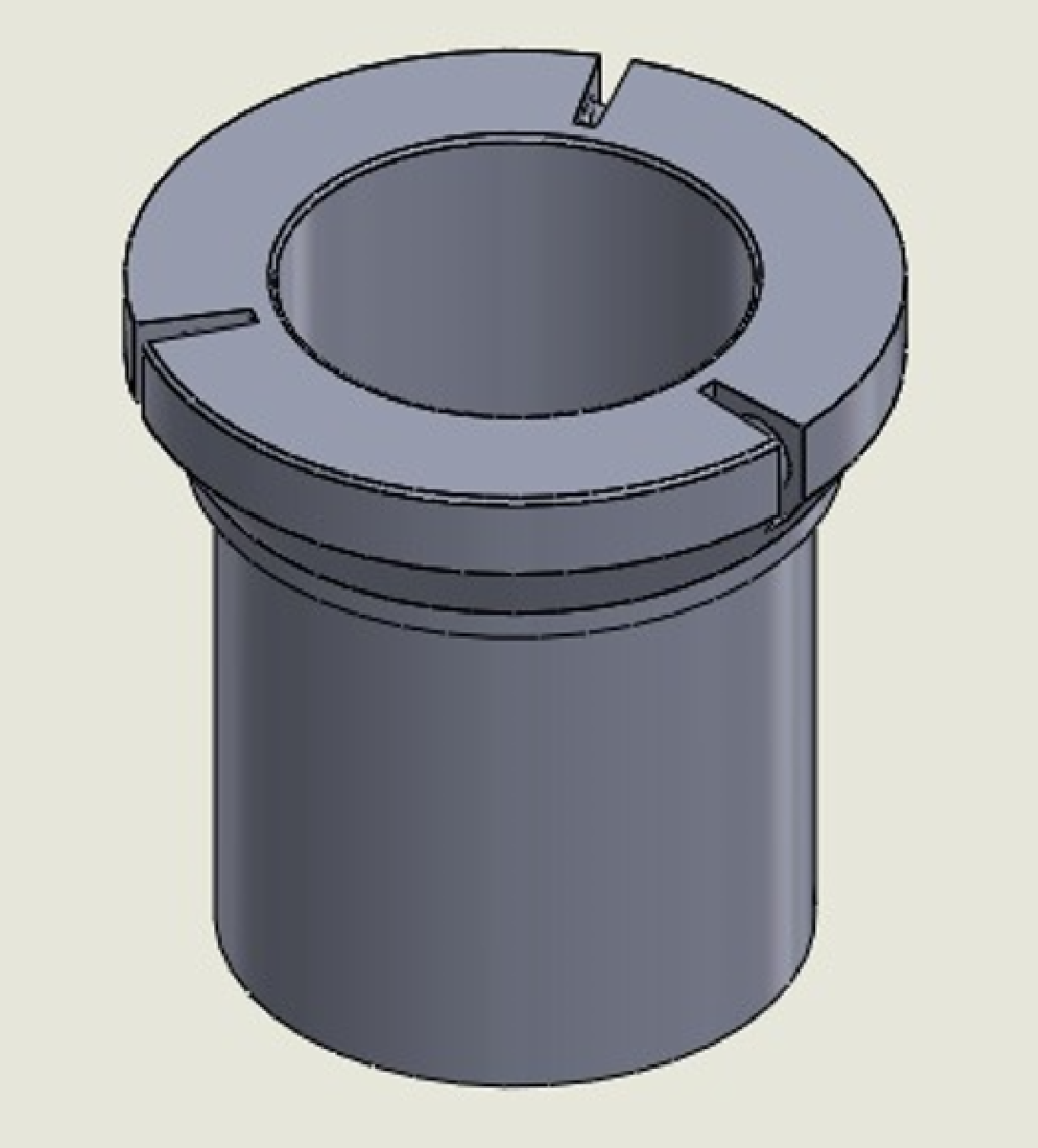}
            {\small\captionof{figure}{Payload A}}
            \vspace{2pt}
        \end{minipage} &
        \begin{minipage}[t]{0.22\textwidth}
            \centering
            \includegraphics[width=\linewidth]{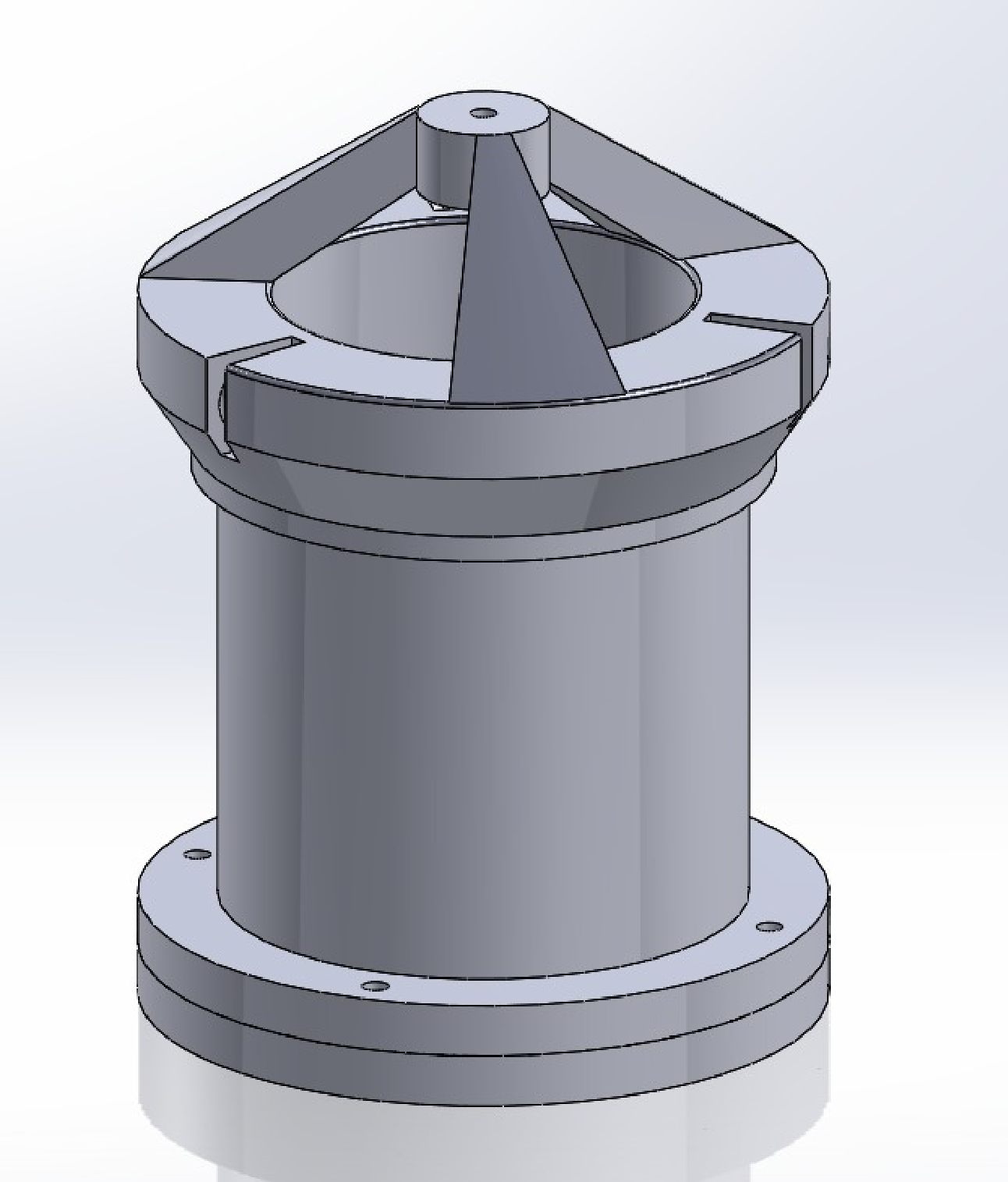}
            {\small\captionof{figure}{Payload B}}
            \vspace{2pt}
        \end{minipage}
    \end{tabular}
\end{figure}

\subsection{Cables}
Cables must be strong enough to withstand the tension induced in them during the operation of the CDPR. While most CDPRs use steel cables, we have used Dyneema cables which have a large strength-to-weight ratio and are easier to bend. In our setup we have used Dyneema cables of 0.9\,mm diameter whose stiffness $k$ is $7\times10^4$\,N/m. We remark that we have chosen these cables since they can easily withstand the tensions obtained from our simulations.

\subsection{Microcontroller}
The stepper motor in our setup is driven by a constant duty cycle ($50\%$) PWM signal whose frequency is varied appropriately to get the desired angular position profile of the motor shaft. To obtain an accurate and smooth motion profile (which will then ensure that the payload moves as desired) the pulses per revolution (PPR) of the motor should be chosen to be large. Larger the PPR, larger is the frequency of the PWM that must be generated by the microcontroller. Furthermore, the microcontroller should be able to command three motors simultaneously. We could not generate three different high frequency PWM signals simultaneously and accurately using Arduino and Raspberry Pi microcontrollers. We were however able to generate these complex signals reliably using NI hardware and so we use it to operate our experimental setup.

%%%%%%%%%%%%%%%%%%%%%%%%%%%%%%%%%%%%%%%%%%%%%%%%%%%%%%%%%%%%%%%%%%%%%
\vspace{-3mm}$$\label{fi2}\includegraphics[scale=0.3]{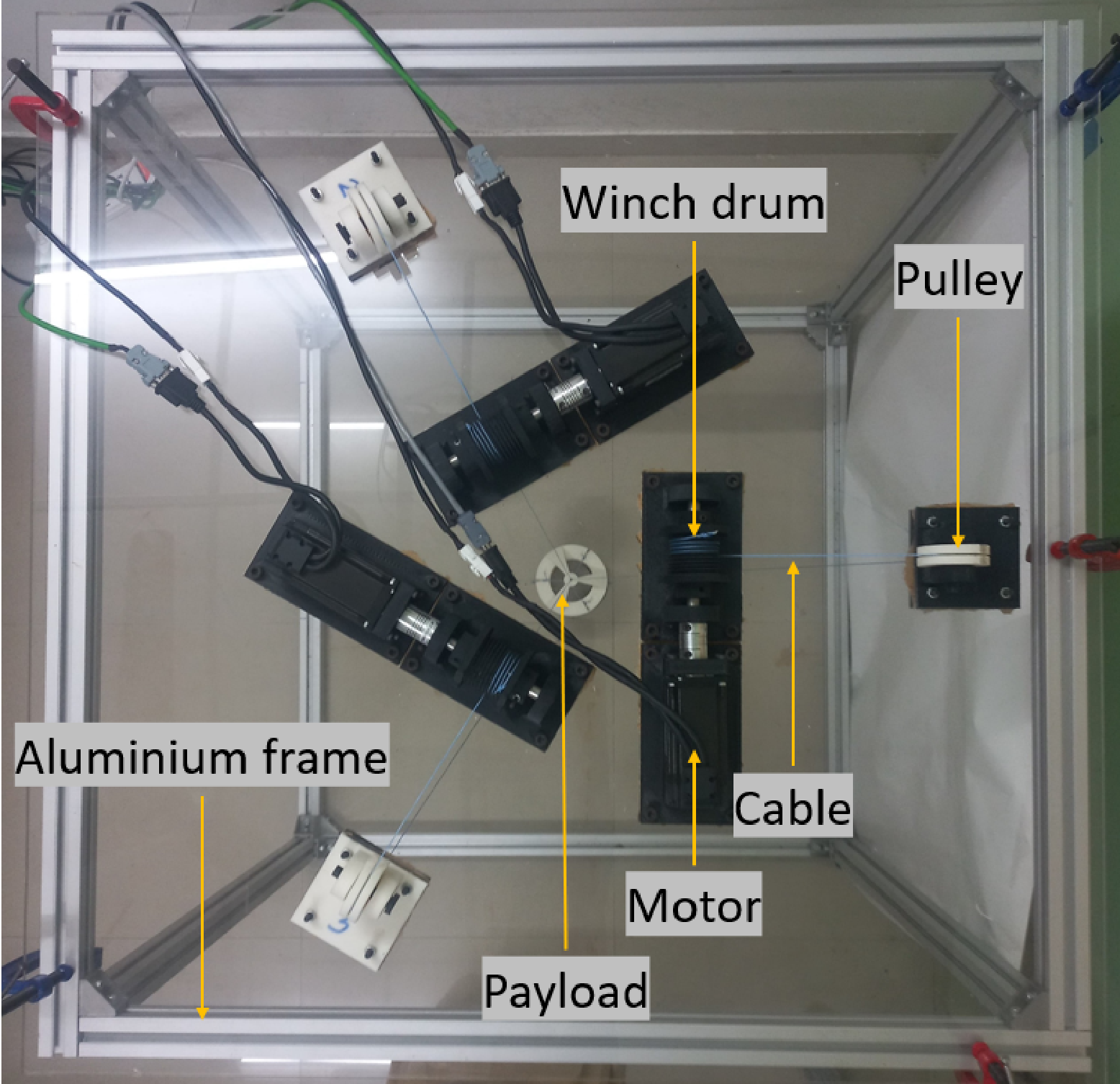} $$
\centerline{\parbox{3.3in}{\vspace{0mm}\centering
\small{Fig. 5. The above picture shows the top view of our experimental setup (CDPR with three cables). The main components are labelled.  }}}\vspace{0.5mm}
%%%%%%%%%%**********%%%%%%%%%%**********%%%%%%%%%%**********%%%%%%%%%%

%%%%%%%%%%%%%%%%%%%%%%%%%%%%%%%%%%%%%%%%%%%%%%%%%%%%%%%%%%%%%%%%%%%%%%
$$\label{fi2}\includegraphics[scale=0.142]{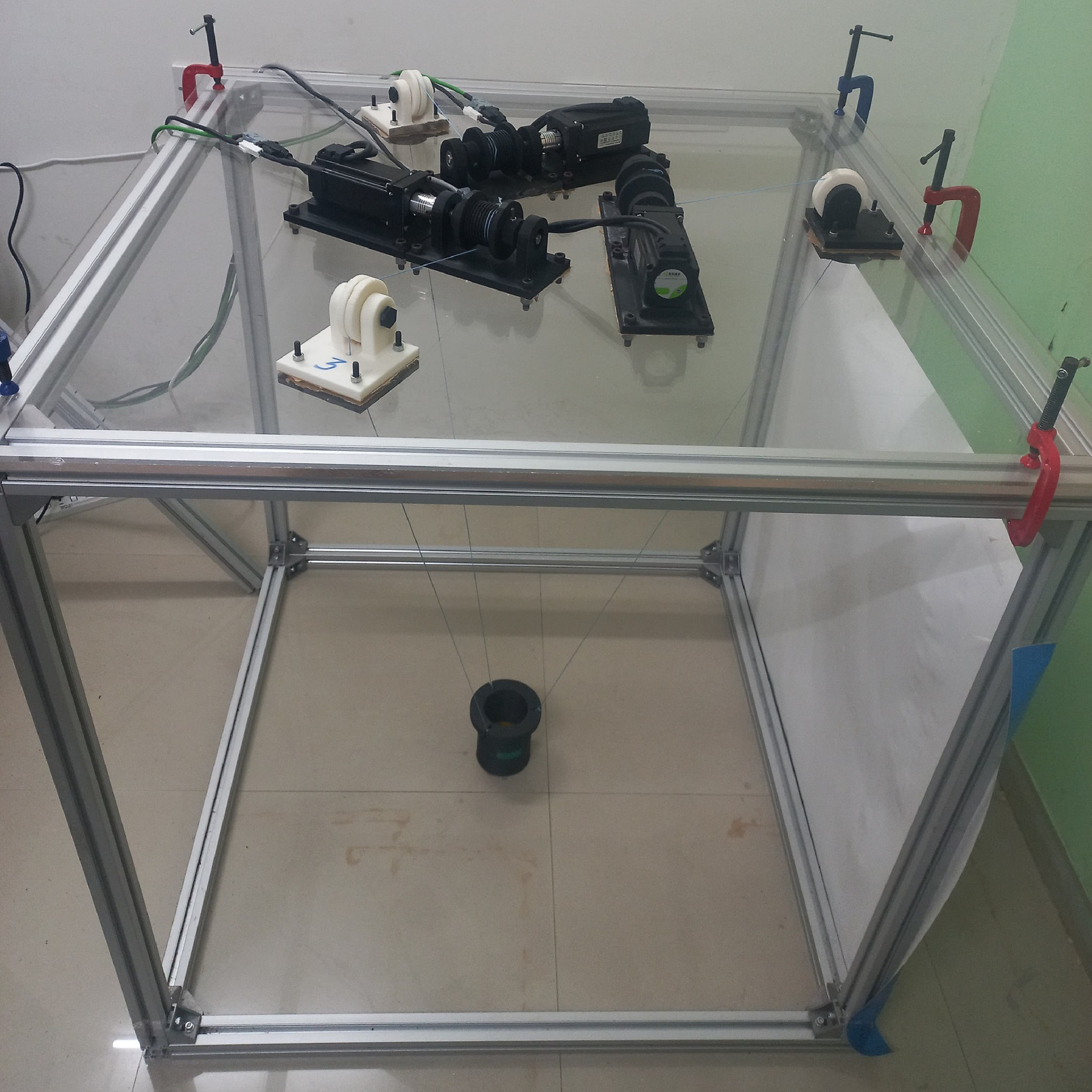} $$
\centerline{\parbox{3.3in}{\vspace{0mm}\centering
\small{Fig. 6. The above picture shows the side view of our experimental setup (CDPR with three cables).}}}\vspace{1mm}
%%%%%%%%%%**********%%%%%%%%%%**********%%%%%%%%%%**********%%%%%%%%%%

\subsection{CAD modeling and fabrication of solid parts}
Many of the solid parts (such as winch drums and pulleys) were modeled in Solidworks and then 3D printed  using ABS filaments. While designing the components, the tolerance of the 3D printer and the bearings should be taken into account since our setup contains tight fit joints.

\subsection{Assembly of CDPR}
The skeleton of our experimental setup is a 1m$\times$1m$\times$1m cubic frame built using 40mm$\times$40mm T-slot aluminium extrusion profiles. These profiles can be connected to each other using cast corner angle brackets, which not only ensure ease of assembly and disassembly, but also render the frame lightweight and robust. The components of the CDPR (including the three motors, winch drums and pulleys) are mounted in a triangular arrangement on an acrylic sheet which is placed on top of the aluminium cubic frame. Figures 5 and 6 show the top view and side view of the CDPR.

\subsection{Operation of the CDPR}
Given a desired trajectory for the payload, the first step is to calculate the desired lengths of the cables (as a function of time). This is done by solving a set of algebraic equations. Then this length requirement is converted to a desired angular velocity profile for the motor shaft and these angular velocity profiles are realized by generating appropriate PWM signals using the microcontroller.

%%%%%%%%%%**********%%%%%%%%%%**********%%%%%%%%%%**********%%%%%%%%%%
%%%%%%%%%%**********%%%%%%%%%%**********%%%%%%%%%%**********%%%%%%%%%%
\section{Experimental results} \label{sec4} %Section 4

We are particularly interested in the synthesis of open-loop control laws which enable the execution of rapid maneuvers in CDPRs. In this section, we consider two simple (rapid) maneuvers, a horizontal maneuver and a vertical maneuver, and design control laws for executing them. We have implemented these control laws on our experimental setup and have verified that they perform satisfactorily.

\subsection{Maneuver 1: Vertical motion of Payload A}
Recall the Payload A shown in Figure 3. In our experiment, the total mass of this payload is 1.5\,kgs. The control objective in our experiment is to transfer the payload from an initial rest position on the vertical axis which is equidistant from all the actuators to a final rest position which is 0.5\,m vertically above the initial position, see Figure 7. We wish to complete this transfer within 1 second. We have considered two open-loop control laws for executing this transfer:\\[1ex]
(i) Constant angular velocity control: In this control law, the angular velocity of each of the motor shafts is kept constant for 1 second, where the constant is chosen so that the payload traverses 0.5\,m in 1 second. This is a discontinuous control law and as expected results in severe oscillations of the payload when implemented.\\[1ex]
(ii) Sinusoidal control law: In this control law, the acceleration profile $a(t)$ for the payload is chosen to be
\begin{equation}\label{eq:acc}
 a(t) = \begin{cases}
  -2\cos 4\pi t + 2 & \text{if } 0 \leq t \leq \frac{1}{2}, \\
  2\cos 4\pi t - 2 & \text{if } \frac{1}{2} < t \leq 1.
\end{cases}
\end{equation}
This profile corresponds to a smooth position profile for the payload which starts at the initial rest position and ends at the final rest position. From this position profile, we have deduced the corresponding angular velocity profile for the motor shaft and implemented it on our setup. This control law ensures a smooth transfer of the payload as desired.

The videos of the experiments which implement the above two control laws can be found here: https://youtu.be/hFJ7xGPeiIM

%%%%%%%%%%%%%%%%%%%%%%%%%%%%%%%%%%%%%%%%%%%%%%%%%%%%%%%%%%%%%%%%%%%%%%
$$\label{fi2}\includegraphics[scale=0.32]{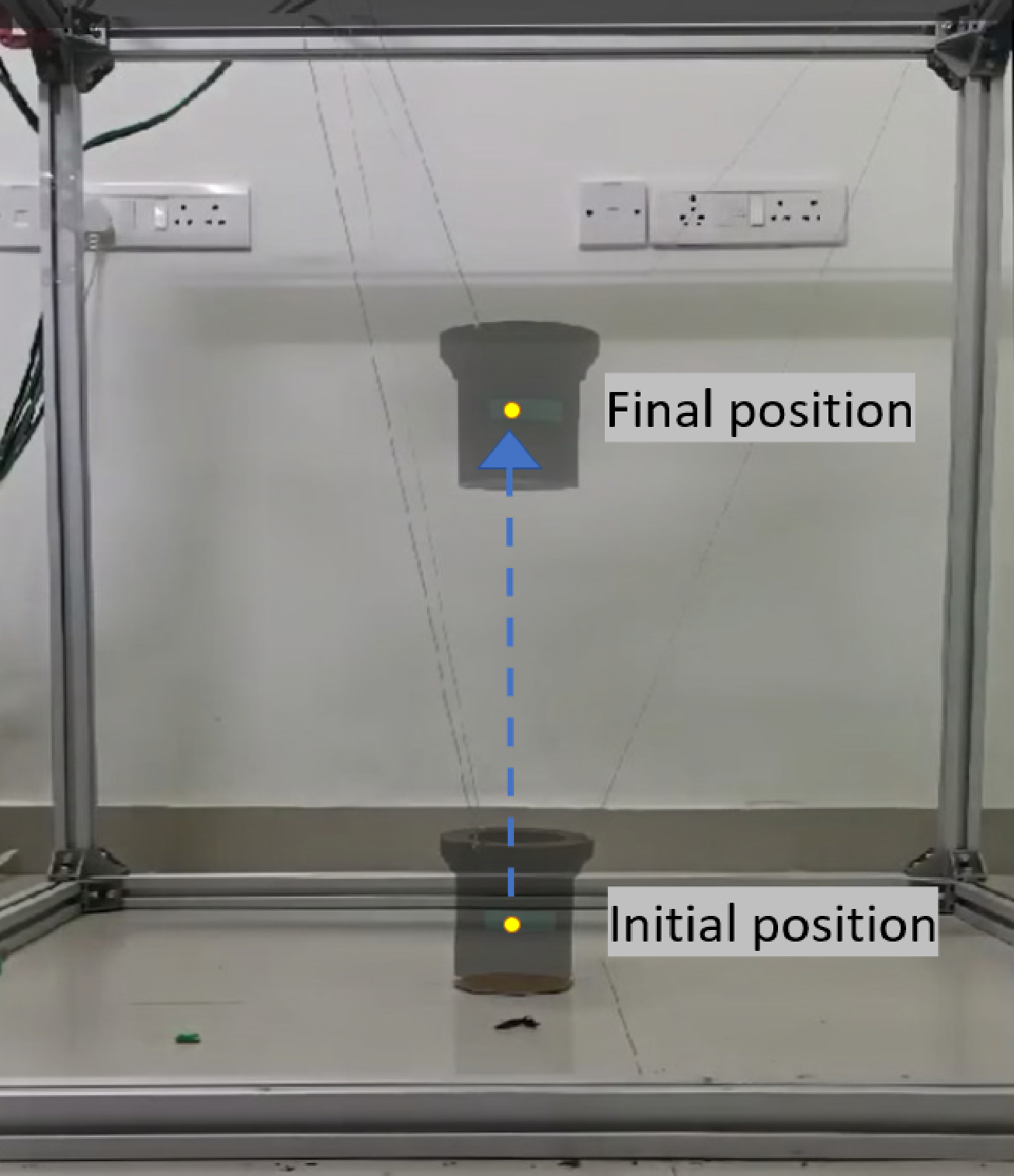} $$
\centerline{\parbox{3.3in}{\vspace{0mm}\centering
\small{Fig. 7. The above picture shows the vertical maneuver to be performed by Payload A. The video of the experiments can be found here: https://youtu.be/hFJ7xGPeiIM}}}\vspace{1mm}
%%%%%%%%%%**********%%%%%%%%%%**********%%%%%%%%%%**********%%%%%%%%%%

\subsection{Maneuver 2: Horizontal motion of Payload B}
Recall the Payload B shown in Figure 4. In our experiment, the total mass of this payload is 2.7\,kgs. The control objective in our experiment is to transfer the payload from an initial rest position on the vertical axis which is equidistant from all the actuators to a final rest position which is 0.3\,m away from the initial position in the same horizontal plane, see Figure 8. We wish to complete this transfer within 2 seconds. We have considered two open-loop control laws for executing this transfer:\\[1ex]
(i) Constant linear velocity control: In this control law, the linear velocity of the anchor point on the payload is kept constant at 0.15 m/s (in the horizontal direction) for 2 seconds. This is a discontinuous control law and as expected results in severe oscillations of the payload when implemented. \\[1ex]
(ii) Pendulum-based control law: The oscillations observed while implementing the previous control law is due to the inertia of the payload which makes it behave like a pendulum suspended about the anchor point. The mathematical model for the payload oscillation is as follows:
\begin{equation}\label{eq:pendulum}
  ml\ddot{x}cos\theta + ml^2\ddot{\theta} + mglsin\theta = 0,
\end{equation}
where $m=2.7$\,kgs, $l=12.9$\,cms, $g=9.8$\,$\textrm{m/s}^2$, $x$ is the position of the anchor point and $\theta$ is the angle made by the payload with the vertical. For small $\theta$, \eqref{eq:pendulum} can be approximated to be
\begin{equation}\label{eq:approx}
  ml\ddot{x} + ml^2\ddot{\theta} + mgl\theta = 0.
\end{equation}
Following \cite{pendulum_trajectory}, we specify the trajectory for $\theta$ as follows:
\begin{equation}\label{eq:theta_des}
 \theta(t) = -2p\sin(\pi t) + p\sin(2\pi t) \FORALL t\in[0,2].
\end{equation}
Here $p= 0.2 \pi/g $ is chosen so that the position profile $x$ for the anchor point obtained by solving \eqref{eq:approx} corresponds to the desired horizontal maneuver. Since $\theta(2)=\dot\theta(2)=0$, the payload also does not oscillate at the end of the maneuver. From the position profile $x$, we have deduced the corresponding angular velocity profile for the motor shaft and implemented it on our setup. This control law ensures a smooth transfer of the payload as desired.

The videos of the experiments which implement the above two control laws can be found here: https://youtu.be/ShxrmmJh6Zw

%%%%%%%%%%%%%%%%%%%%%%%%%%%%%%%%%%%%%%%%%%%%%%%%%%%%%%%%%%%%%%%%%%%%%%
$$\label{fi2}\includegraphics[scale=0.32]{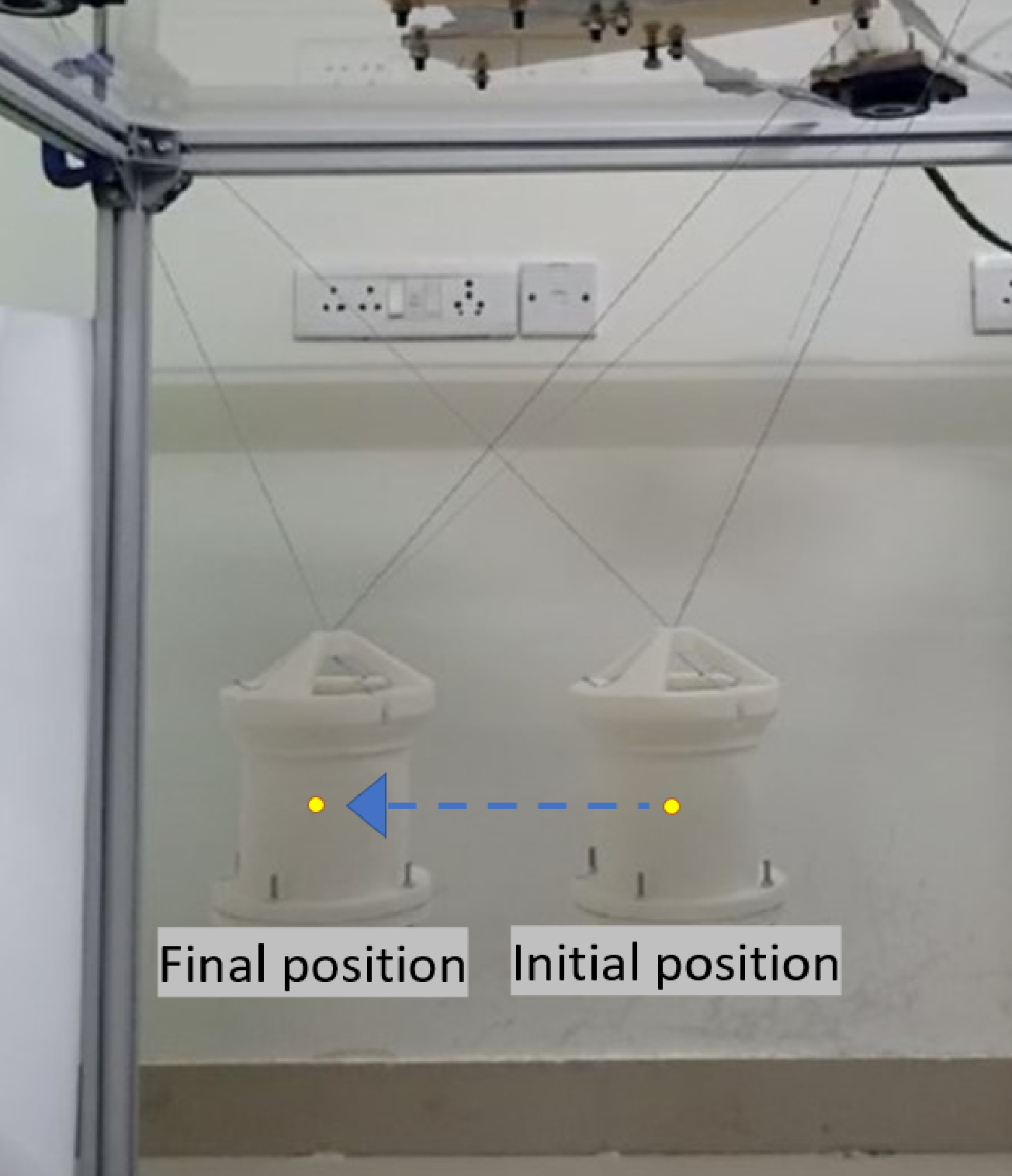} $$
\centerline{\parbox{3.3in}{\vspace{0mm}\centering
\small{Fig. 8. The above picture shows the horizontal maneuver to be performed by Payload B. The video of the experiments can be found here: https://youtu.be/ShxrmmJh6Zw}}}\vspace{3mm}
%%%%%%%%%%**********%%%%%%%%%%**********%%%%%%%%%%**********%%%%%%%%%%

Future work will focus on designing control algorithms for executing more complex maneuvers and implementing these algorithms on our experimental setup.

%%%%%%%%%%**********%%%%%%%%%%**********%%%%%%%%%%**********%%%%%%%%%%
%%%%%%%%%%**********%%%%%%%%%%**********%%%%%%%%%%**********%%%%%%%%%%

\end{document}